\documentclass[conference,10pt]{IEEEtran}

\makeatother

\usepackage{verbatim}
\usepackage{varioref}
\usepackage{acro} 
\usepackage{amsmath}
\usepackage{amssymb}
\usepackage{graphicx}
\usepackage{fixmath}
\usepackage{url}
\usepackage{algorithm}
\usepackage{algpseudocode}
\usepackage{color}
\usepackage{xargs}                      
\usepackage[pdftex,dvipsnames]{xcolor}  
\usepackage[colorinlistoftodos,prependcaption,textsize=tiny]{todonotes}
\usepackage{etoolbox}
\usepackage{subcaption}
\usepackage{todonotes}
\usepackage[inline]{enumitem}
\makeatletter
\patchcmd{\Ginclude@eps}{"#1"}{#1}{}{}
\makeatother

\usepackage{multicol}
\usepackage{amssymb}

\usepackage{cite}

\newcommand{\mbsf}[1]{\boldsymbol{\mathsf{#1}}}

\newcommand{\rv}{\mathbf{r}}

\newcommand{\vv}{\mathbf{v}}
\newcommand{\wv}{\mathbf{w}}
\newcommand{\xv}{\mathbf{x}}
\newcommand{\yv}{\mathbf{y}}
\newcommand{\zv}{\mathbf{z}}

\newcommand{\Av}{\mathbf{A}}
\newcommand{\Bv}{\mathbf{B}}

\newcommand{\Iv}{\mathbf{I}}

\newcommand{\rs}{\mathsf{r}}

\newcommand{\xs}{\mathsf{x}}

\newcommand{\wvs}{\mbsf{w}}
\newcommand{\xvs}{\mbsf{x}}
\newcommand{\yvs}{\mbsf{y}}

\DeclareAcronym{cs}{
	short = CS,
	long  = compressed sensing
}
\DeclareAcronym{awgn}{
	short = AWGN,
	long  = additive white Gaussian noise
}
\DeclareAcronym{snr}{
	short = SNR,
	long  = signal-to-noise ratio
}
\DeclareAcronym{se}{
	short = SE,
	long  = state evolution
}
\DeclareAcronym{amp}{
	short = AMP,
	long  = approximate message passing
}
\DeclareAcronym{lamp}{
	short = LAMP,
	long  = learned AMP
}
\DeclareAcronym{oamp}{
	short = OAMP,
	long  = orthogonal AMP
}
\DeclareAcronym{alamp}{
	short = ALAMP,
	long  = analytical learned AMP
}
\DeclareAcronym{lgmamp}{
	short = L-GM-AMP,
	long  = learned Gaussian-mixture AMP
}
\DeclareAcronym{bamp}{
	short = BAMP,
	long  = Bayesian approximate message passing
}
\DeclareAcronym{gamp}{
	short = GAMP,
	long  = generalized approximate message passing
}
\DeclareAcronym{lista}{
	short = LISTA,
	long  = learned ISTA
}
\DeclareAcronym{ista}{
	short = ISTA,
	long  = iterative soft thresholding algorithm
}
\DeclareAcronym{pdf}{
	short = pdf,
	long  = probability density function
}
\DeclareAcronym{cdf}{
	short = cdf,
	long  = conditional density function
}
\DeclareAcronym{ls}{
	short = LS,
	long  = least squares
}
\DeclareAcronym{mse}{
	short = MSE,
	long  = mean squared error
}
\DeclareAcronym{mmse}{
	short = MMSE,
	long  = minimum mean squared error
}
\DeclareAcronym{nmse}{
	short = NMSE,
	long  = normalized mean squared error
}
\DeclareAcronym{ser}{
	short = SER,
	long  = section error rate
}
\DeclareAcronym{iid}{
	short = i.i.d.,
	long  = independent and identically distributed
}
\DeclareAcronym{bg}{
	short = BG,
	long  = Bernoulli-Gauss
}
\DeclareAcronym{ssc}{
	short = SSC,
	long  = Sparse Superposition Codes
}
\DeclareAcronym{ml}{
	short = ML,
	long  = Machine Learning
}
\DeclareAcronym{gm}{
	short = GM,
	long  = Gaussian-mixture
}
\DeclareAcronym{pme}{
	short = PME,
	long  = posterior-mean estimator
}
\DeclareAcronym{bs}{
	short = BS,
	long  = block-sparsity
}
\DeclareAcronym{bsm}{
	short = BSM,
	long  = block sparsity model
}
\DeclareAcronym{bswsbm}{
	short = BSWSBM,
	long  = block sparsity with sparse blocks model
}
\DeclareAcronym{dps}{
	short = DPS,
	long  = delay power spectrum
}
\DeclareAcronym{mmv}{
	short = MMV,
	long  = multiple measurement vectors
}
\DeclareAcronym{fft}{
	short = FFT,
	long  = Fast Fourier Transformation
}
\DeclareAcronym{ifft}{
	short = IFFT,
	long  = Inverse Fast Fourier Transformation
}
\DeclareAcronym{kl}{
	short = KL,
	long  = Kullback-Leibler
}
\DeclareAcronym{mc}{
	short = MC,
	long  = Monte-Carlo
}
\DeclareAcronym{lasso}{
	short = LASSO,
	long  = least absolute shrinkage and selection operator
}

\newcommandx{\unsure}[2][1=]{\todo[linecolor=red,backgroundcolor=red!25,bordercolor=red,#1]{#2}}

\usepackage{enumitem}

\floatname{algorithm}{Algorithm}
\algnewcommand\algorithmicinput{\textbf{Input:}}
\algnewcommand\algorithmicinit{\textbf{Initialization:}}
\algnewcommand\INPUT{\item[\algorithmicinput]}
\algnewcommand\INIT{\item[\algorithmicinit]}
\algnewcommand\algorithmicoutput{\textbf{Output:}}
\algnewcommand\OUTPUT{\item[\algorithmicoutput]}

\usepackage[bottom]{footmisc}

\begin{document}

\title{Plug-And-Play Learned Gaussian-mixture Approximate Message Passing}


\author{$\text{Osman Musa}$, $\text{Peter Jung}$ and $\text{Giuseppe Caire}$  \\
Communications and Information Theory Group, Technische Universit{\"a}t Berlin\\
Email: \{osman.musa,peter.jung,caire\}@tu-berlin.de.}
\maketitle

\begin{abstract}
Deep unfolding showed to be a very successful approach for accelerating and tuning classical signal processing algorithms.
In this paper, we propose \ac{lgmamp} - a plug-and-play \ac{cs} recovery algorithm suitable for any i.i.d. source prior. 
Our algorithm builds upon Borgerding's \ac{lamp}, yet significantly improves it by adopting a universal denoising function within the algorithm. 
The robust and flexible denoiser is a byproduct of modelling source prior with a \ac{gm}, which can well approximate continuous, discrete, as well as mixture distributions.
Its parameters are learned using standard backpropagation algorithm.
To demonstrate robustness of the proposed algorithm, we conduct \ac{mc} simulations for both mixture and discrete distributions.
Numerical evaluation shows that the \ac{lgmamp} algorithm achieves state-of-the-art performance without any knowledge of the source prior. 
\end{abstract}



\acresetall
\section{Introduction}

We consider the problem of recovering compressible signals embedded in a high-dimensional data space from low-dimensional representations. More specifically, we consider recovering compressible $N$-dimensional vector $\xv$ from $m<N$ linear and noisy measurements arranged in a vector $\yv$, observed using a measurement matrix $\Av$, i.e., find  $\xv$ from
\begin{equation}
\label{eq:cs_problem}
\yv = \Av\xv + \wv,
\end{equation}
where $\wv$ is \ac{iid} additive noise. This problem, known as noisy \ac{cs} recovery problem \cite{donoho2006compressed,candes2006robust}, received a lot of attention in the last two decades. As a result, many algorithms were proposed to solve (\ref{eq:cs_problem}), and an overview of those can be found in \cite{maleki2013asymptotic,marques2019areview}. Even though, many of these algorithms offer provable recovery guarantees in the presence of noise, their parameters need to be tuned in often not a straightforward way \cite{maleki2010optimally}.

As an alternative to the classical signal processing approach, recent trend was to combine iterative recovery algorithms with tools from \ac{ml}. Here, a prominent idea is to unfold an iterative algorithm into a deep neural network and, using training data, learn (i.e., optimize) network parameters in a procedure that has objective to minimize a loss function (e.g., \ac{nmse}) \cite{balatsoukasstimming2019deep,monga2020algorithm}. For example, the authors in \cite{karol2010learning} unfold \ac{ista} and learn its parameters from training data by minimizing reconstruction \ac{mse}. The resulting algorithm, called \ac{lista}, requires significantly less iterations to achieve the same prediction error as \ac{ista}. 
Next good candidate for unfolding is the \ac{amp} algorithm\cite{donoho2010message,donoho2010messageII}, whose behaviour, for sub-Gaussian \ac{iid} matrices, and in the large system limit, is predicted by the \ac{se} \cite{bayati2011dynamics}. 
In \cite{borgerding2017amp}, Borgerding et al. propose \ac{lamp} algorithm, which builds upon \ac{amp} and uses one of several parametric family of component-wise denoisers.
The empirical results \cite{borgerding2017amp} show that, by learning per layer filter weights, as well as the parameters of the denoiser function, the \ac{lamp} network significantly improves upon both \ac{lista} from \cite{karol2010learning} and \ac{amp} algorithm from \cite{donoho2010message,donoho2010messageII}.
Even though the \ac{lamp} network shows excellent empirical performances, it is not clear, however, apart from empirically investigating the \ac{nmse}, whether the chosen parametric family of denoisers is good for a given source prior.


In this work, inspired by \cite{borgerding2017amp}, we, therefore, examine a general-purpose denoiser within the \ac{lamp} algorithm. 
Instead of defining a parametric family of denoisers, we consider modelling the source signal prior as an \ac{iid} \ac{gm} distribution, which is a byproduct of modelling the source signal prior as an \ac{iid} \ac{gm} distribution. 
Later, just as in \ac{bamp}, we adopt the "optimal" denoiser function that would minimize \ac{nmse} had the assumed \ac{gm} prior match the true unknown prior. The parameters of the \ac{gm}, which are at the same time parameters of the denoiser function, as well as the filter weights, are learned in the same way as in \cite{borgerding2017amp}, i.e., by unfolding the \ac{amp} algorithm and training the network on training data samples. The \ac{lamp} algorithm with assumed \ac{gm} prior-based denoising function will be called \ac{lgmamp}. 

Using Gaussian distribution as the prior's building blocks has several advantages, namely:
\begin{itemize}
\item the resulting denoiser function $\eta(\cdot)$, and it's derivative $\frac{d}{dr} \eta(\cdot)$ can be calculated analytically,
\item if the overall objective is to minimize reconstruction \ac{nmse}, a good approximation of a discrete component in the source prior is a Gaussian distribution with matching mean and very small variance, and
\item a Gaussian mixture can model a variety of continuous distribution.
\end{itemize}

The idea of modelling the non-zero part of a signal prior with a \ac{gm} within the \ac{amp} framework has been investigated in \ac{cs} in \cite{borgerding2017amp}. There, the authors optimize the parameters of the mixture using expectation maximization step. A similar approach of tuning parameters of an assumed model distribution using the Method-of-Moments was proposed in \cite{goertz2017fast}. Both approaches  learn the parameters of the prior within the iterations of the algorithms, and show solid empirical performances. Both tuning procedures are built within the classical \ac{amp} algorithm, and, therefore, do not benefit in terms of accuracy and speed of convergence from learning the weight matrix. On the other hand, in \ac{lgmamp}, by learning the filter weights, we potentially increase the convergence speed of the \ac{amp} algorithm, while keeping the flexibility and robustness offered by the \ac{gm}-based denoiser. However, with our approach we loose the connection to the \ac{se}, and theoretical guarantees for the recovery are still open.

\subsection*{Notation} 
Vectors and matrices are represented by boldface characters. 
A Gaussian \ac{pdf} with mean $\mu$ and variance $\sigma^2$ is denoted by $\mathcal{N}(\cdot;\,\mu, \sigma^2)$. 
Unless otherwise specified $\|\cdot\|$ corresponds to the Euclidian (or $\ell_2$) norm. 
Random variables, random vectors, and random matrices are denoted by sans-serif font, e.g. $\sf a$, $\mbsf{a}$, and $\mbsf{A}$, respectively. 
$[N]$ denotes the set of positive integers up to $N$, i.e., $[N] = \{1,\ldots,N\}$. 
All zeros vector of size $N$ is denoted with $\mathbf 0_N$.
The $N \times N$ identity matrix is denoted with $\Iv_{N}$. 

\section{Learned Gaussian-mixture AMP}
It was shown in  \cite{borgerding2017amp} that the $t$-th layer of the tied \ac{lamp} network can be written as  
\begin{align}
\rv^{(t)} &= \xv^{(t-1)} +  \Bv\zv^{(t)},\\
\xv^{(t+1)} &= \eta \big(  \rv^{(t)} ; \,{\sigma^2}^{(t)}, \mathrm{\Theta}^{(t)} \big),\\
 b^{(t)} &= \frac{1}{m}  \sum_{i=1}^N \frac{\partial [\eta \big( \rv^{(t)}, {\sigma^2}^{(t)}, \mathrm{\Theta}^{(t)} \big)]_i}{\partial r_i }\\
 \zv^{(t)} &= \yv - \Av \xv^{(t-1)}  + b^{(t-1)} \zv^{(t-1)}, \label{eq:Onsager}
\end{align}
where ${\sigma^2}^{(t)}$ is the effective noise variance at the $t$-th layer, which can be estimated as $\sigma^{(t)} = \| \zv^{(t)}\|_2 / \sqrt{m}$, $\mathrm{\Theta}^{(t)}$ is the vector of denoiser parameters at that layer, and $\Bv$ is the learned weight (filter) matrix. 
Since the presence of the Onsager term in (\ref{eq:Onsager}) allows for the decoupled measurement model \cite{montanari2012graphical}, i.e., makes $\rv^{(t)}$ to be distributed as $\xv + \vv^{(t)}$, where $\vv^{(t)} \sim \mathcal N(\mathbf 0_N, \, {\sigma^2}^{(t)}  \Iv_{N})$, the \ac{mmse} estimator is given by 
\begin{equation}
\label{eq:eta}
\eta(r^{(t)}_n; \,\sigma^{(t)}, \mathrm{\Theta}^{(t)}) = \mathbb E [\xs_n \thinspace \vert \thinspace \rs^{(t)}_n = r^{(t)}_n ; \,\sigma^{(t)}, \mathrm{\Theta}^{(t)}].
\end{equation}
For a prior given with \ac{gm} distribution:
\begin{equation}
\label{eq:nGauss}
p(x_n \,;\,\mathrm{\Theta}_{\text{\ac{gm}}}) = \sum_{l=1}^L \omega_l \thinspace \mathcal N (x_n; \mu_l, \sigma_{l}^2),
\end{equation}
where $ \sum_{l=1}^L \omega_l = 1$, $0\leq \omega_l \leq 1$ for all $l \in [L]$, and $\mathrm{\Theta}_{\text{\ac{gm}}} = \cup_{l=1}^L \{\omega_{l}, \mu_{l},  \sigma_{l}^2\}$, the conditional \ac{pdf} of $\xs_n$ given  $\rs^{(t)}_n$ can be written as 
\begin{equation}
\label{px_n|r_n}
p(x_n \thinspace \vert \thinspace r_n ; \, \sigma^2, \mathrm{\Theta}_{\text{\ac{gm}}}) =  \sum_{l=1}^L {\bar\beta_{n,l}}  \thinspace \mathcal N(x_n; \gamma_{n,l}, \nu_{n,l}), 
\end{equation}
where
\begin{equation}
\label{gamma_nu}
\begin{split}
\bar\beta_{n,l} &=  \frac{ \beta_{n,l}}{\sum_{l'=1}^L \beta_{n,l'}},\\
\beta_{n,l} &= \omega_l \thinspace  \mathcal N(r_n; \mu_l, \sigma_{l}^2 + \sigma^2),\\
\gamma_{n,l} &= \frac{r_n\sigma_{l}^2 + \mu_l\sigma^2}{\sigma_{l}^2 + \sigma^2},\\
\nu_{n,l} &= \frac{\sigma_{l}^2 \sigma^2}{\sigma_{l}^2 + \sigma^2}.
\end{split}
\end{equation}
It should be noted that, in order to simplify notation, in (\ref{px_n|r_n}) and (\ref{gamma_nu}) we leave out the iteration number $^{(t)}$ in both $\rs^{(t)}_n$, and $\sigma^{(t)}$. Using (\ref{px_n|r_n}), the \ac{mmse} estimator $\eta(r^{(t)}_n; \,\sigma^{(t)}, \mathrm{\Theta}^{(t)}_{\text{\ac{gm}}})$ from (\ref{eq:eta}), and it's derivative with respect to $r_n$ are given with
\begin{equation}
\begin{split}
\eta(r_n; \, \sigma^2, \mathrm{\Theta}_{\text{\ac{gm}}}) &= \sum_{l=1}^L {\bar\beta_{n,l}} \thinspace \gamma_{n,l},\\
\frac{d}{dr_n}\eta(r_n;\, \sigma^2, \mathrm{\Theta}_{\text{\ac{gm}}}) &=  \sum_{l=1}^L \frac{\sigma_{l}^2 }{\sigma_{l}^2 + \sigma^2} {\bar\beta_{n,l}} + \gamma_{n,l}  \frac{d \bar\beta_{n,l}}{dr_n},\\
\end{split}
\end{equation}
where
\begin{equation}
\frac{d \bar\beta_{n,l}}{dr_n} = \sum_{l'=1}^L  \bar\beta_{n,l} \thinspace\bar\beta_{n,l'}  \Big( \frac{r_n-\mu_{l'}}{\sigma_{l'}^2 + \sigma^2} -   \frac{r_n-\mu_l}{\sigma_{l}^2 + \sigma^2}\Big).\\
\end{equation}

\subsection{Learning the \ac{lgmamp} Parameters}
Training of the \ac{lgmamp} network with $T_{\text{max}}$ layers, which has $Nm + 3LT_{\text{max}}$ tunable parameters ($\{\Bv, \cup_{t=1}^{T_{\text{max}}} \mathrm{\Theta}_{\text{\ac{gm}}}^{(t)}\}$),  is described in Algorithm \ref{alg:tied_learning}. 
There, we start by initializing parameters of a \ac{gm} $\mathrm{\Theta}_{\text{\ac{gm}}}^{(0)}$, followed by learning the weight matrix $\Bv$. 
Later, in contrast to \cite[Algorithm 2]{borgerding2017amp}, to reduce the training resources, in each iteration $t$ we initialize new layer's parameters $\mathrm{\Theta}_{\text{\ac{gm}}}^{(t)}$ with last learned layer's \ac{gm} parameters and learn only the new layer's parameters. 
Finally, we refine parameters of all layers up to and including the new layer using smaller learning rates. 

It should be noted that it is also possible to tie parameters $\mathrm{\Theta}_t$ across layers $t$, but this showed not to bring a significant reduction in training resources and to deteriorate the gain in \ac{nmse}.
Similarly, it is possible to untie the filter matrix $\Bv$, and, therefore, for each layer have a layer specific $\Bv^{(t)}$. However, this approached showed not to bring a significant benefit in \ac{nmse}.

\begin{algorithm}
	\caption{Tied LAMP-GMP parameter learning}{ }
	\begin{algorithmic}[1]
		\INIT $\Bv = \Av^T, \mathrm{\Theta}_{\text{\ac{gm}}}^{(0)} = \cup_{l=1}^L \{\omega_{l,0}, \mu_{l,0},  \sigma_{l,0}^2\}$
		\State Learn $\Bv$        
		\For{$t\in[1,T]$}
			\State Initialize $\mathrm{\Theta}_{\text{\ac{gm}}}^{(t)}  = \mathrm{\Theta}_{\text{\ac{gm}}}^{(t-1)} $ 
			\State Learn $\mathrm{\Theta}_{\text{\ac{gm}}}^{(t)}$ with fixed $\mathrm{\Theta}_{\text{tied}}^{(t-1)}$ \label{alg:tied_learning_step_learn} 
			\State Refine $\mathrm{\Theta}_{\text{tied}}^{(t)}=\{\Bv, \cup_{t'=1}^t \mathrm{\Theta}_{\text{\ac{gm}}}^{(t')}\}$ 
		\EndFor
	\end{algorithmic}
	\label{alg:tied_learning}
\end{algorithm}

\section{Numerical Results}
In the numerical evaluation of the proposed algorithm, the entries of the sensing matrix $\Av$ are drawn once independently from a zero-mean Gaussian distribution with variance $1/m$, and kept fixed.
To compare performances of the \ac{lgmamp} algorithm to the \ac{amp} and \ac{lamp} algorithms with matched prior, the per-iteration \ac{nmse}, given with
\begin{equation}
\label{eq:nmse}
\text{\ac{nmse}}^{(t)} =  \mathbb E\big[ \| \hat\xvs^{(t)} - \xvs\|_2^2 \, \big\vert \, \Av \big] / \, \mathbb E \big[ \| \xvs\|_2^2 \big],
\end{equation}
is used as the performance metric. As the expectation over the source and noise vectors $\xv$ and $\wv$ in (\ref{eq:nmse}) is difficult to evaluate, we run a \ac{mc} simulation, and compute average of 10000 realizations of the empirical per-iteration $\text{\ac{nmse}}^{(t)}\triangleq \| \hat\xv^{(t)} - \xv\|_2^2 /  \| \xv\|_2^2 $, averaged over $\xv$ and $\wv$. 

\begin{figure}[t]
	\centering
	\includegraphics[width=0.42\textwidth]{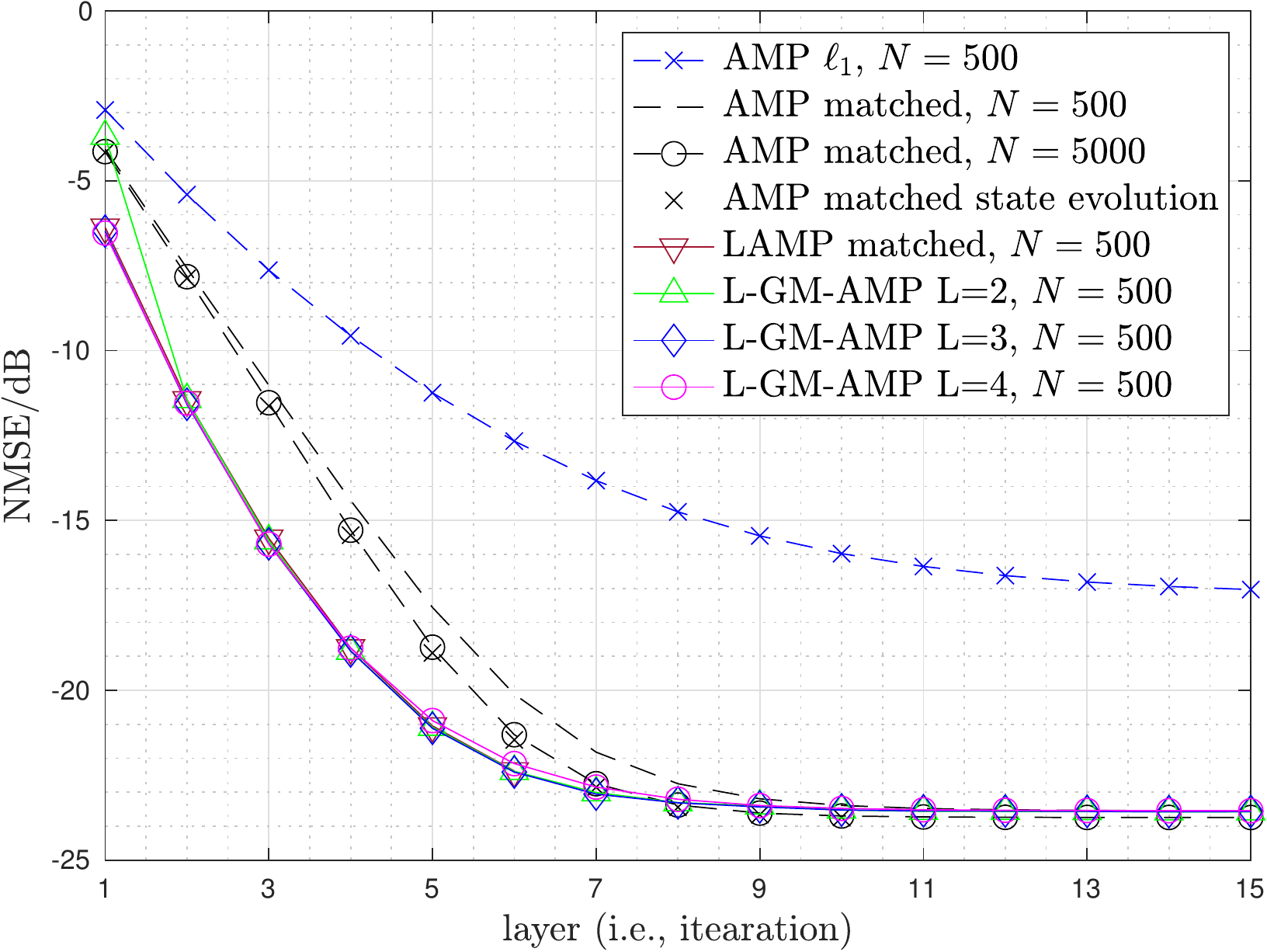}
	\caption{\ac{nmse} of \ac{amp}, matched \ac{lamp} and \ac{lgmamp} against layer number for a \ac{bg} mixture prior. Parameters: $\delta=0.5$, $\epsilon=0.1$, $\text{\ac{snr}}=20 \text{dB}$.}
	\label{fig:bg_prob_nmse}
\end{figure}

Given the measurement radio $\delta$ and sparsity $\epsilon$, in each instance of the \ac{mc} simulation, we take $m=\delta N$ measurements of a $500$-long $\epsilon N$-sparse source vector (i.e., $N = 500$).
Given an \ac{snr}, defined as 
\begin{equation}
\label{eq:snr_def}
\text{\ac{snr}} = \mathbb E\big[\Vert \yvs \Vert^2\big] / \, \mathbb E \big[\Vert \wvs \Vert^2\big] = \frac{\epsilon}{\delta} \frac{\sigma_x^2}{\sigma_w^2},
\end{equation}
where $\sigma_x^2$ is the variance of the nonzero entries of $\xv$, the entries of the noise vectors are drawn independently from a zero-mean Gaussian distribution with variance $\sigma_w^2=\text{\ac{snr}} \, \frac{\epsilon}{\delta} \, \sigma_x^2$.

Training procedure, given in Algorithm \ref{alg:tied_learning},  that was kindly made available online\footnote{\url{http://github.com/mborgerding/onsager_deep_learning}} by authors of \cite{borgerding2017amp}, was implemented in Python using open-source library TensorFlow 1.14. The network is trained using Adam optimizer \cite{kingma2014adam}, with the training rate of  $10^{-3}$ (step \ref{alg:tied_learning_step_learn}, Alg. \ref{alg:tied_learning}), and with training rates of $10^{-4}$, and $10^{-5}$ for fine tuning in step 5. For training, we use mini-batch size of $10^{3}$, while for validation and testing the batch size is $10^{4}$.
It should be noted that we used 64-bit double-precision floating-point data type to solve possible numerical issues.

To show generality of the proposed approach we compare its empirical reconstruction \ac{nmse} for different \ac{iid} prior distributions $p_{\xs}$. Specifically, we consider:
\begin{itemize}
\item a \acl{bg} mixture distribution, i.e., 
\begin{equation}
p_{\xs}(x) = (1-\epsilon) \, \delta(x) + \epsilon \, \mathcal N (x; 0, \sigma_w^2),
\end{equation}
where $\epsilon$ is probability of a nonzero entry, and $\sigma_w^2$ is power of the Gaussian component.
\item a discrete distribution over alphabet $\mathcal A=\{\varepsilon_1, \dots, \varepsilon_L\}$ of size $L$, i.e., 
$p_{\xs}(x) = \sum_{l=1}^{L} w_l \, \delta(x-\varepsilon_l)$,
where $w_l$ is probability of $\xs$ being equal to $\varepsilon_l$.
\end{itemize}

\subsection{Results}

\begin{figure}[t]
\centering
	\includegraphics[width=0.42\textwidth]{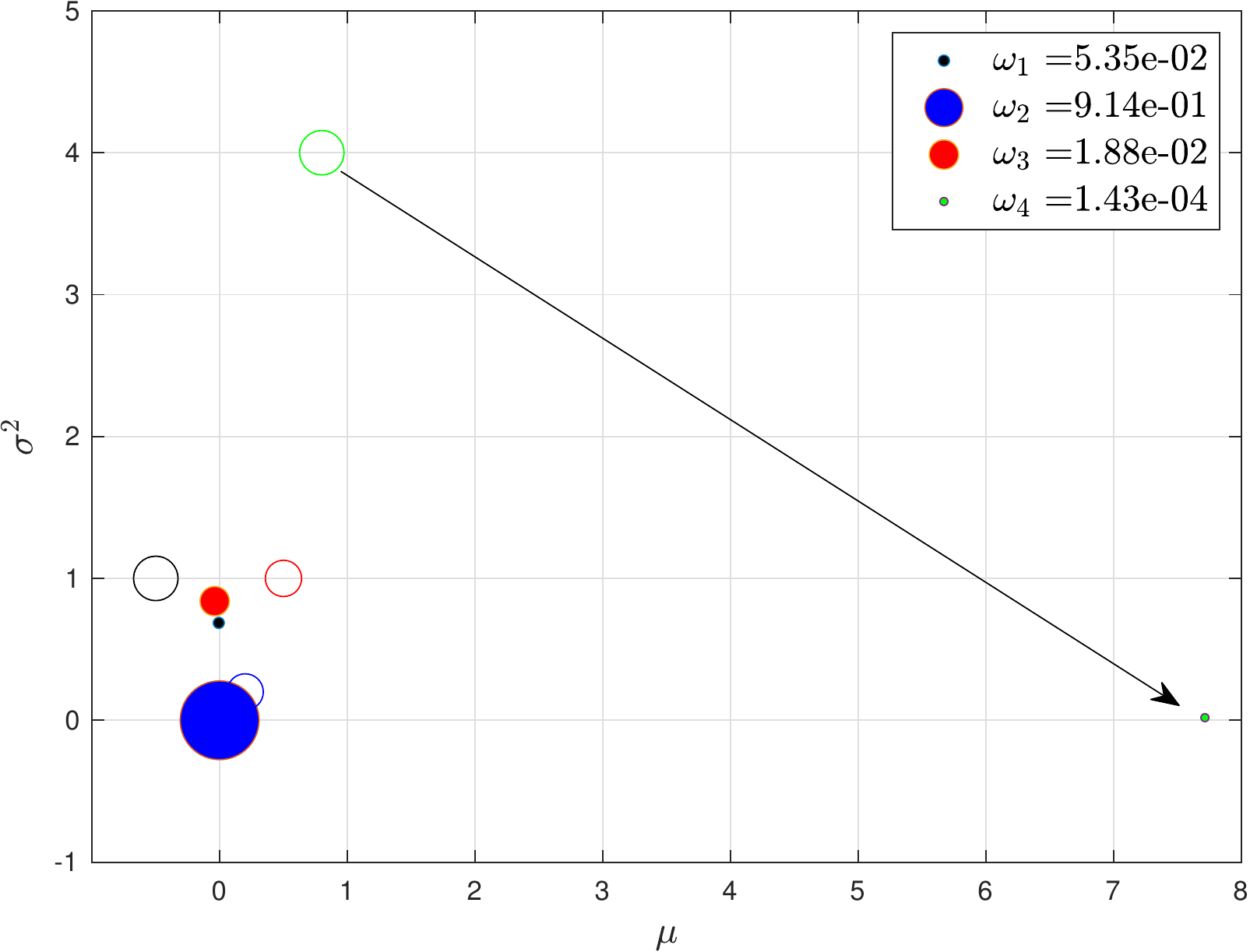}
	\caption{Learned prior parameters (full circles) of the \ac{gm} with $4$ components at the 9th layer, and their initial values (empty circles). The true prior is a \ac{bg} mixture with $\epsilon=0.1$.}
	\label{fig:bg_prob_thetas}
\end{figure}

\subsubsection{\ac{bg} mixture prior} For a \ac{bg} mixture prior with $\epsilon=0.1$ and $\sigma_x^2=1$, Figure \ref{fig:bg_prob_nmse} shows the \ac{nmse} against iterations of the classical (i.e., untrained) \ac{amp} and the \ac{lamp} algorithms that use different priors (i.e., denoisers). 
Namely, we consider the so-called $\ell_1$ prior, named after the $\lambda \| \xv \|_1$ term appearing in the objective function of \ac{lasso}, and  the matched prior, which assumes perfect knowledge of the source distribution. 
Furthermore, we show results for \ac{lgmamp} with $L=2$, $L=3$, and $L=4$. 
First, we can see that the \ac{amp} $\ell_1$ algorithm is inferior to the matched \ac{amp}, which is a consequence of $\ell_1$ prior not approximating well enough the true source prior.
Second, we observe a certain discrepancy, especially at the first few iterations, between matched \ac{amp} for a problem with $N=500$, and the same algorithm for a 10 times larger problem (i.e., $N=5000$), for which the \ac{nmse} is well predicted by the \ac{se}. 
We conclude that the concentration of measure phenomenon, on which \ac{amp} was built, takes full effect for larger $N$.
Finally, we can see that \ac{lgmamp} matches the performance of \ac{lamp} with matched denoiser, and does not suffer from over-parametrization, ever when $L=4$. 

Figure \ref{fig:bg_prob_thetas} shows the initial parameters as well as learned parameters of a \ac{gm} with 4 components at the 9th layer. Initial parameters are indicated by empty circles while the learned prior parameters are indicated by full circles. The size (surface) of a circle is proportional to the weight (i.e., probability) $\omega_l$ of the respective component in the mixture. 
Observing the blue circle, we see that, since $\omega_2 = 9.14\mathrm{e}{-01}$ and the variance close to zero, the network has learned that the source distribution has a significant density centred around zero. Furthermore, our network has detected two Gaussian components (red and black circles) with zero mean and variance close to true variance of the Gaussian component of the \ac{bg} mixture, with different weights. This indicate that one way the network handles possible over-paramtrization is to split one Gaussian into two with Gaussian the same mean and variance, whose combined weight makes the weight of the initial Gaussian component. Another way the network handles possible over-paramtrization is shown with the fourth component (green circle). This "unnecessary" component, whose learning trajectory was indicated by the arrow, was given a vary large mean, a small variance, and insignificantly small weight $\omega_4 = 1.43\mathrm{e}{-04}$, diminishing its contribution in for the denoiser.

\begin{figure}[t]
	\centering
	\includegraphics[width=0.42\textwidth]{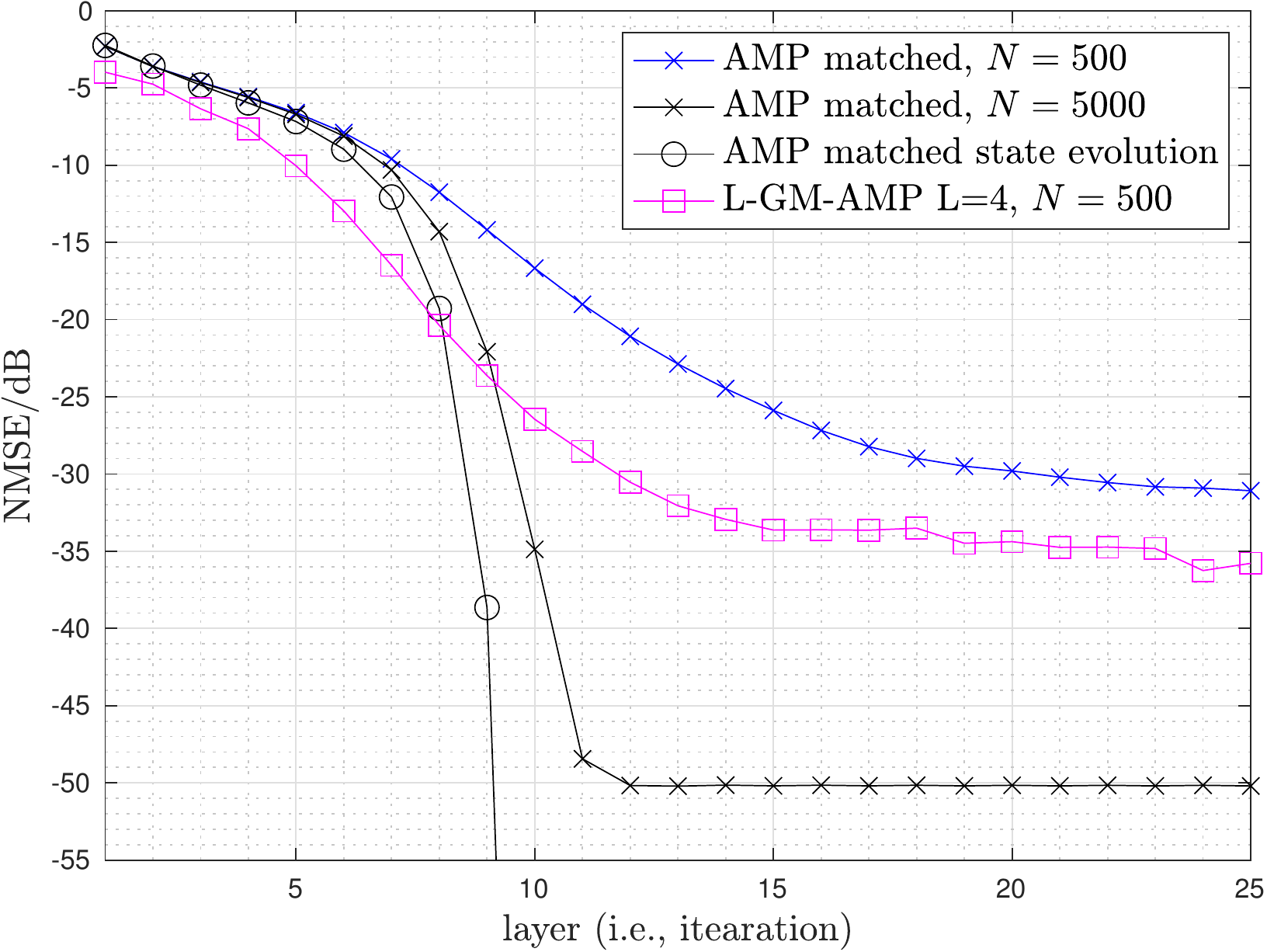}
	\caption{\ac{nmse} of \ac{amp} and \ac{lgmamp} against layer number for a binary prior. Parameters: $\delta=0.65$, $\text{\ac{snr}}=15 \text{dB}$, $p_{\xs}(-1)=p_{\xs}(+1)=0.5$. }
	\label{fig:bpsk_prob_nmse}
\end{figure}

\subsubsection{Discrete distribution prior}
Next we consider a symmetric Bernoulli distribution with alphabet $\{ -1,+1\}$, which is used to model an anti-sparse source that maps input binary symbols to BPSK (binary antipodal) transmit symbols. 
Since matched \ac{lamp} network showed to be unstable for training, in Figure \ref{fig:bpsk_prob_nmse} we show results only for the matched \ac{amp} and the \ac{lgmamp} algorithm
For the considered prior, by comparing the \ac{nmse} of matched \ac{amp} algorithms evaluated on different problem sizes ($N = 500$ and $N = 5000$) with its \ac{se} prediction, we observe that the algorithm is heavily influenced by finite dimensions of the problem, even more than in the \ac{bg} prior case. 
We, therefore, conclude that the concentration of measure phenomenon is influenced not only by the problem size, but also by the source distribution.
Some of the loss caused by finite dimensions of the problem, is, however, reduced by the \ac{lgmamp} network. We conjecture that this is a result of learned weight matrix decoupling the measurements, which \ac{amp} algorithm fails to do for moderate-size problems (e.g., $N = 500$).

\begin{figure}[t]
	\centering
	\includegraphics[width=0.42\textwidth]{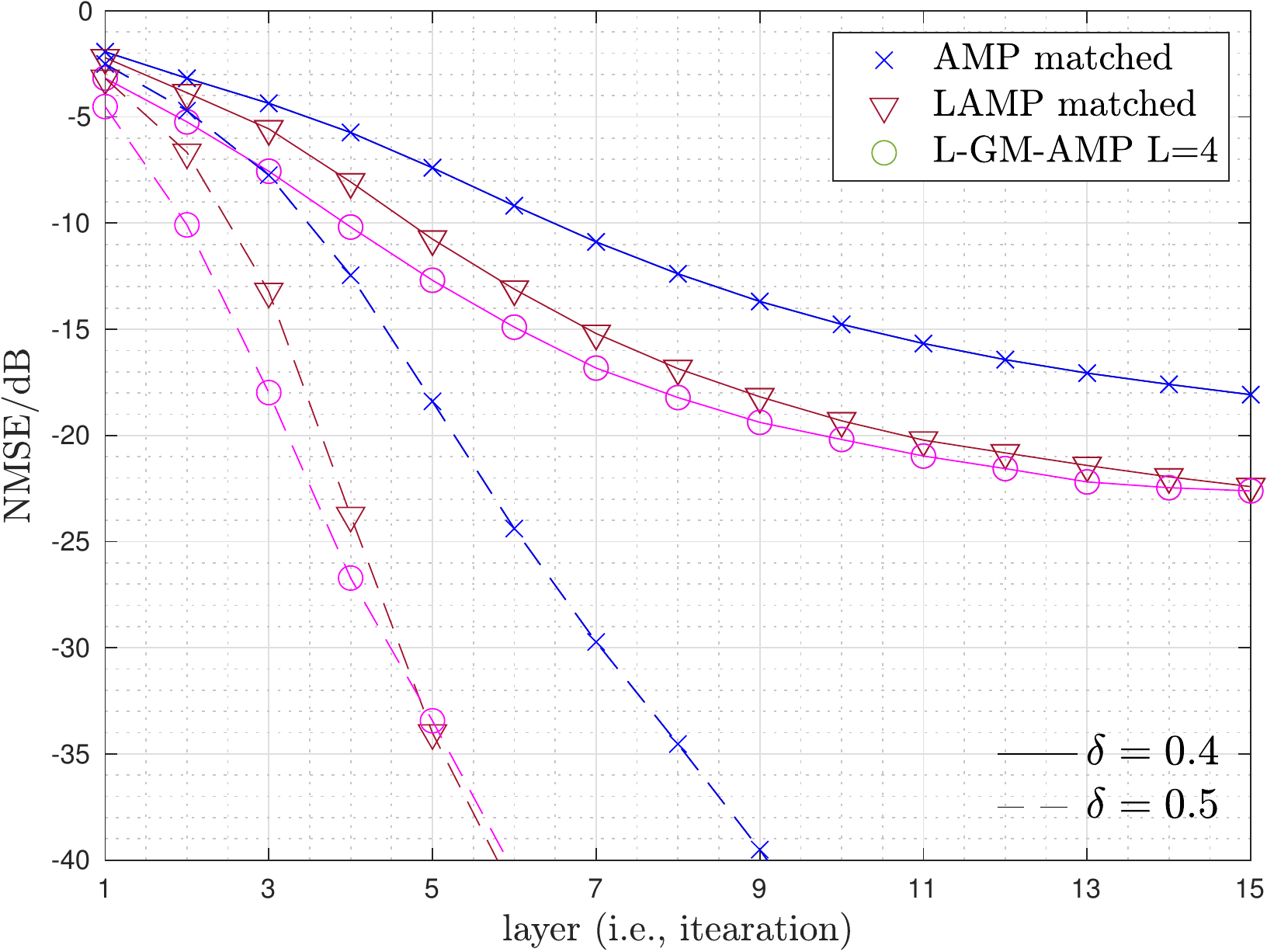}
	\caption{\ac{nmse} of \ac{amp}, matched \ac{lamp} and \ac{lgmamp} against layer number for a discrete prior. Parameters: $\text{\ac{snr}}=20 \text{dB}$, $N = 500$, $p_{\xs}(-1)=p_{\xs}(+1)=0.05$, $p_{\xs}(0)=0.9$. }
	\label{fig:dm_prob_nmse}
\end{figure}

Figure \ref{fig:dm_prob_nmse} shows the \ac{nmse} against iterations of the matched \ac{amp}, the matched \ac{lamp} and the \ac{lgmamp} algorithm for a sparse discrete distribution over alphabet $\{ -1,0,+1\}$, with respective probabilities 0.05, 0.9, and 0.05. Two measurement rates are considered, namely $\delta=0.4$ for which the \ac{nmse} values are shown with solid lines, and $\delta=0.5$ for which the \ac{nmse} values are shown with dashed lines. 
In both cases, at first few iterations, the \ac{lgmamp} algorithm outperforms the matched \ac{amp} algorithm, and even slightly the matched \ac{lamp} algorithm. 
The gain is, however, expected to vanish with the increasing iteration number. 

\section{Conclusions}
In this paper we presented \ac{lgmamp} for recovering unknown sparse as well as anit-sparse vectors from noisy \ac{cs} measurements.  Although reminiscent of Borgerding's \ac{lamp} \cite{borgerding2017amp}, it differs in the adoption of a universal plug and play denoising function. The robust and flexible denoiser is based on modelling source prior with a \ac{gm}, which can well approximate continuous, discrete, as well as mixture distributions. The parameters of the assumed prior, and at the same time of the denoising function, are learned using standard backpropagation algorithm. Numerical results show that the \ac{lgmamp} algorithm achieves state-of-the-art performance offered by (L)\ac{amp} with perfect knowledge of the source prior. 
Extending this work for recovering signals from noisy non-linear \ac{cs} measurements remains an interesting problem for future work.


\bibliographystyle{IEEEtran} \bibliography{references}

\end{document}